\documentclass[letterpaper, 10 pt, conference]{ieeeconf}  
\IEEEoverridecommandlockouts                              
\usepackage{graphics} 
\usepackage{epsfig} 
\usepackage{mathptmx} 
\usepackage{times} 
\usepackage{amsmath} 
\usepackage{amssymb}  

\title{\LARGE \bf User profile-driven large-scale multi-agent learning from demonstration in federated human-robot collaborative environments}
\author{Georgios Th. Papadopoulos$^{1}$, Asterios Leonidis$^{1}$, Margherita Antona$^{1}$ and Constantine Stephanidis$^{1}$$^{2}$\\\{gepapado,leonidis,antona,cs\}$@$ics.forth.gr
\thanks{$^{1}$Institute of Computer Science (ICS), Foundation for Research and Technology - Hellas (FORTH), GR-70013, Heraklion, Crete, Greece}
\thanks{$^{2}$Computer Science Department (CSD), University of Crete (UoC), GR-70013, Heraklion, Crete, Greece}
}

\begin{document}

\maketitle
\thispagestyle{empty}
\pagestyle{empty}

\begin{abstract}
Learning from Demonstration (LfD) has been established as the dominant paradigm for efficiently transferring skills from human teachers to robots. In this context, the Federated Learning (FL) conceptualization has very recently been introduced for developing large-scale human-robot collaborative environments, targeting to robustly address, among others, the critical challenges of multi-agent learning and long-term autonomy. In the current work, the latter scheme is further extended and enhanced, by designing and integrating a novel user profile formulation for providing a fine-grained representation of the exhibited human behavior, adopting a Deep Learning (DL)-based formalism. In particular, a hierarchically organized set of key information sources is considered, including: a) User attributes (e.g. demographic, anthropomorphic, educational, etc.), b) User state (e.g. fatigue detection, stress detection, emotion recognition, etc.) and c) Psychophysiological measurements (e.g.  gaze, electrodermal activity, heart rate, etc.) related data. Then, a combination of Long Short-Term Memory (LSTM) and stacked autoencoders, with appropriately defined neural network architectures, is employed for the modelling step. The overall designed scheme enables both short- and long-term analysis/interpretation of the human behavior (as observed during the feedback capturing sessions), so as to adaptively adjust the importance of the collected feedback samples when aggregating information originating from the same and different human teachers, respectively.
\end{abstract}

\section{Introduction}
\label{sec:Introduction}

{\it{Learning from Demonstration}} (LfD) constitutes the most dominant paradigm for robot programming through direct interaction with humans \cite{Argall09}. It relies on the fundamental principle of robots acquiring new skills by learning to imitate a (human) teacher \cite{Ravichandar20}. LfD has so far been successfully associated with multiple aspects of robotics technology, including human-robot interaction, machine learning, machine vision and motor control \cite{furuta2020motion}\cite{seleem2020development}. The main advantageous characteristics that have led to the widespread adoption of LfD in multiple and diverse application cases are \cite{Billard16}: a) It enables robot programming by non-expert users, b) It allows time-efficient learning, c) It enables adaptive robotic behaviors and d) It facilitates the robots to operate in complex and time-varying environments.

With respect to the different types of human demonstration means considered, LfD approaches generally fall into three main categories \cite{Ravichandar20}:
\begin{itemize}
\item Kinesthetic teaching, where the human teacher provides demonstrations by physically moving the robot through the desired motions \cite{Caccavale19};
\item Teleoperation, where an external input (e.g. joystick, graphical user interface or other means) is used to guide the robot \cite{Zhang18};
\item Passive observation, where the robot passively observes the exhibited user behavior for learning \cite{Liu18}.
\end{itemize}

A critical factor for the success of any LfD scheme, while also in close relation with the particularities of the selected application case, concerns the adopted methodology for refining the robot learned policies. Different types of approaches have been introduced so far, including: a) Reinforcement learning, where a trial and error methodology is adopted to learn a policy to solve a given problem \cite{Zhang16}, b) Optimization, where the optimal solution is searched based on given criteria \cite{Cheng15}, c) Transfer learning, where knowledge of a task or a domain is used to enhance the learning procedure for another task \cite{Brys15}, d) Apprenticeship learning, where the desired performance is modeled through the use of a set of demonstrated samples that serve as a template \cite{Wulfmeier15}, e) Active learning, where the robot decides when to ask for an optimal response from a (human) expert to a given state and to use these active samples to improve its policy \cite{Judah12} and f) Structured predictions, where robotic actions are considered as a sequence of dependent predictions \cite{Droniou14}.

Although extensive research efforts have been devoted in the LfD field over the past years, crucial research challenges, namely multi-agent learning \cite{Hussein17} and long-term autonomy \cite{Kunze18}, still remain to be reliably addressed. Achieving the latter will act as a tremendous facilitator for enabling the wide-spread use of robots in large-scale, open and complex environments; this is the typical case when considering e.g. industrial manufacturing scenarios. The above need to be also investigated in conjunction with typical challenges in Human-Robot Interaction (HRI) schemes, like developing appropriate user interfaces, variance in human performance, variability in knowledge across human subjects, learning from noisy/imprecise human input, learning from very large or very sparse datasets, incremental learning, etc.

The current work adopts the Federated Learning (FL) conceptualization that has very recently been introduced for designing large-scale LfD human-robot collaborative environments \cite{papadopoulos2020towards}, aiming at providing reliable solutions, among others, to the current critical challenges of multi-agent learning and long-term autonomy. In particular, the cognitive architecture of \cite{papadopoulos2020towards} is further elaborated and enhanced, by designing and integrating a novel user profile formulation for estimating a fine-grained representation of the exhibited human behavior. The introduced user model follows the Deep Learning (DL)-based formalism; hence, rendering the overall approach end-to-end learnable. Specifically, a set of key information sources that are hierarchically organized is considered, including: a) User attributes concerning each human teacher’s background (e.g. demographic, anthropomorphic, educational, etc.), b) User state information that encodes particular human behavioral characteristics (e.g. fatigue detection, stress detection, emotion recognition, etc.) and c) Psychophysiological measurements that convey critical details about the user's mental state, personality and idiosyncrasy (e.g. gaze, electrodermal activity, heart rate, etc.). Then, a combination of Long Short-Term Memory (LSTM) and stacked autoencoders, with appropriately defined neural network architectures, is employed for the modelling step and for producing a concrete user profile representation. The overall designed scheme supports both short- and long-term analysis/interpretation of the human behavior (as observed during the feedback capturing sessions), so as to adaptively modulate the importance of the collected feedback samples when aggregating information originating from the same and different human teachers, respectively.

The remainder of the paper is organized as follows: Section \ref{sec:OpenIssues} describes challenges and open issues currently present in the LfD learning field. Additionally, Section \ref{sec:FL} outlines the already introduced cognitive architecture for FL-based multi-agent human-robot collaborative learning. Moreover, Section \ref{sec:UserProfile} details the designed DL-based user profile modelling and knowledge aggregation approach. Finally, conclusions are drawn in Section \ref{sec:Conclusions}.

\section{Open issues in LfD learning}
\label{sec:OpenIssues}

LfD has been shown to lead to significant advances in multiple and significantly diverse operational scenarios, involving the use of both manipulator \cite{vogt2017system}\cite{bhattacharjee2019towards} and mobile \cite{pan2017agile}\cite{loquercio2018dronet} robots. Regarding the targeted goal of the overall LfD procedure, this can be categorized to different levels of abstraction \cite{Ravichandar20}, including the learning of: a) Policies, i.e. the estimation of a function that generates the desired behavior, b) Cost and reward functions, where the ideal behavior is considered to stem out from the optimization of a hidden function, c) Plans, i.e. high level structured schemes, composed of several sub-tasks or primitive actions, and d) Multiple outcomes simultaneously, by jointly modeling complex behaviors at multiple levels of abstraction.

As already mentioned in Section \ref{sec:Introduction}, despite the extensive research efforts devoted by the LfD community, key technological challenges and open issues still remain to be robustly addressed, including, among others, the following ones \cite{Ravichandar20}\cite{Hussein17}\cite{Billard16}\cite{Zhu18}:
\begin{itemize}
\item To involve in a more robust way a broad number of teachers with different styles of and possibly conflicting demonstrations; 
\item To transfer skills across multiple agents, including multiple and diverse types of robots;
\item To simultaneously learn multiple complex tasks, while storing and reusing prior knowledge at large scale;
\item To robustly implement incremental learning schemes;
\item To reinforce the generalization ability;
\item To model compound tasks;
\item To implement multi-agent imitation learning schemes;
\item To operate in realistic, dynamic, time-varying and complex environments.
\end{itemize}

Towards providing a reliable solution to the above aspects, a novel cognitive architecture for multi-agent LfD robotic learning has recently been introduced \cite{papadopoulos2020towards}, targeting to enable the efficient deployment of open, scalable and expandable robotic systems
in large-scale and complex environments. In particular, the designed architecture capitalizes on the recent advances in the Artificial Intelligence (AI) (and especially the DL field), by establishing a FL-based framework for incarnating a multi-human multi-robot collaborative learning environment (as will be discussed in Section \ref{sec:FL}), going far beyond other literature approaches investigating relatively straight-forward implementations of FL schemes (without incorporating the human factor in the learning loop) in specific robotic tasks (e.g. autonomous navigation \cite{Liu19lifelong}, motion planning \cite{bretan2019robot}, visual perception \cite{zhou2018real}, etc.).

However, the conventional FL mechanism, which is a purely data-driven scheme and typically employs a simple `Federated Averaging' operator for knowledge aggregation \cite{mcmahan2017communication}, lacks theoretical guarantees for convergence under realistic settings \cite{li2019convergence}, i.e. when non-independent-and-identically-distributed (non-IID) data are available at the various network nodes. The latter assumption is highly likely to hold when considering human demonstration data (as required in the LfD scheme). In this context, the current work elaborates the multi-agent cognitive architecture conceptualization of \cite{papadopoulos2020towards}, by introducing comprehensive DL-grounded user profile modelling and corresponding sophisticated knowledge aggregation strategies (explained in Section \ref{sec:UserProfile}).

\section{FL-based multi-agent human-robot collaborative learning}
\label{sec:FL}

\subsection{Fundamental conceptualization}
\label{ssec:FLConceptualization}

\begin{figure}[t]
\centering
\includegraphics[width=0.45\textwidth]{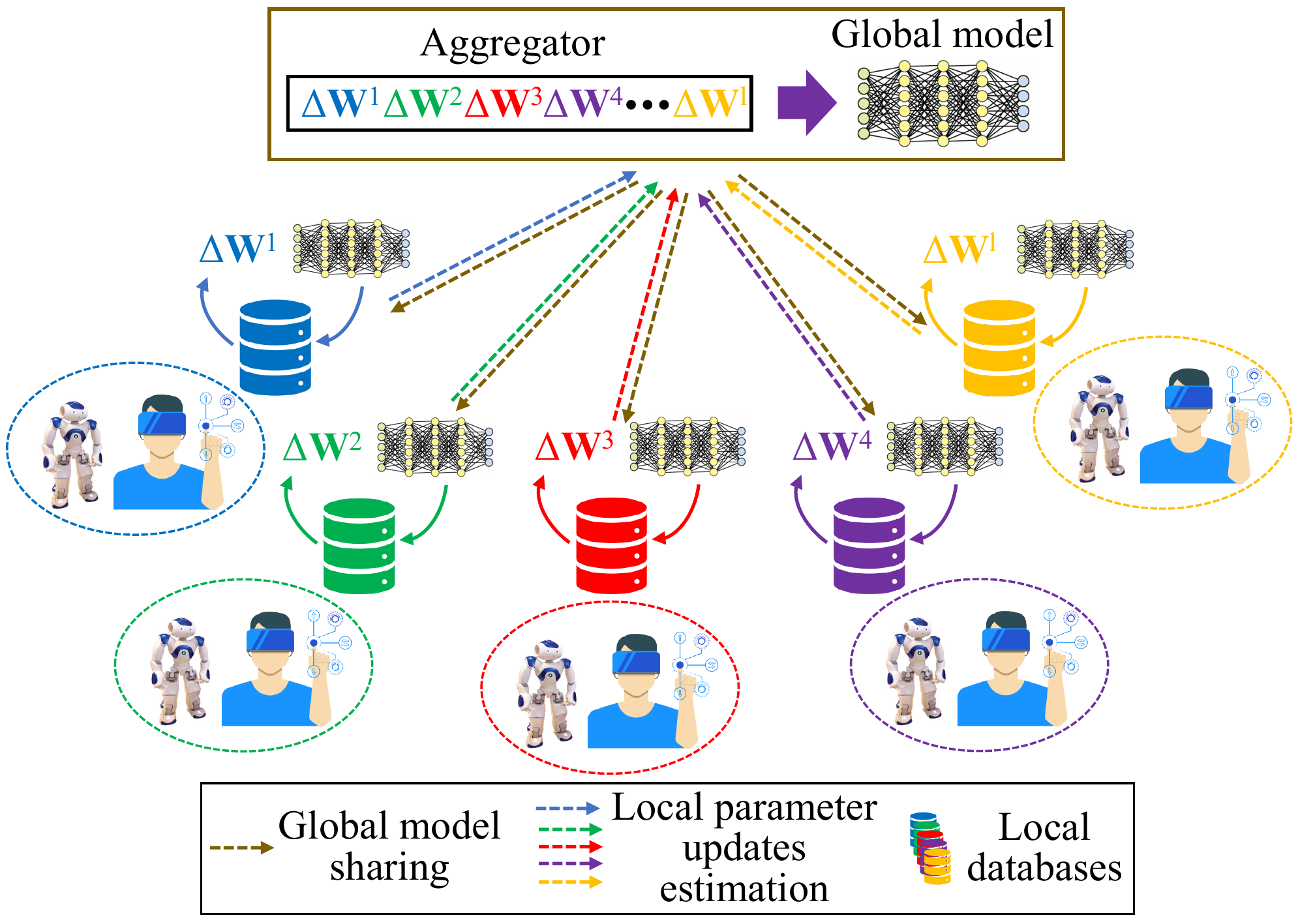}
\caption{High-level representation of the employed cognitive architecture for LfD learning.}
\label{f:FL}
\end{figure}

According to \cite{papadopoulos2020towards}, a multi-agent cognitive architecture for LfD learning is introduced that is grounded on the fundamental mechanism of the FL paradigm, while a high-level representation is provided in Fig. \ref{f:FL}. In particular, a global AI model $\textbf{W}$ (that bears the targeted robot policy to be learned), which is aimed to be collaboratively trained and shared among a network of human-robot pairs, is initially constructed using proxy data (stored either offline or at a central node). Model $\textbf{W}$ materializes a cognitive process \cite{papadopoulos2017CVPR}\cite{papadopoulos2016human} underpinning the robotic task of interest, e.g. sensing, navigation, manipulation, control, human-robot interaction, etc. Subsequently, the model is made available to the network and downloaded by each node (i.e. every employed robot). Every node encapsulates a local database that is used to estimate improved updates of the global model parameters (e.g. using conventional Stochastic Gradient Descent (SGD) for the case of Neural Network (NN)-based AI modules), without making the local (federated) data available to the network (i.e. being a by-definition privacy-aware method). The computed local parameter updates (denoted $\Delta \textbf{W}^l$ in Fig. \ref{f:FL}) are asynchronously transmitted back to the central node (using encrypted communication), where an aggregation mechanism is responsible for combining them and periodically producing a new version of the global model. The overall process is iterative, i.e. continuously estimating improved versions of the global model. It needs to be highlighted that each local computational node can maintain a customized version of the global model, while using the locally stored data for fine-tuning purposes. Moreover, in the current work a single robot policy (unique global model $\textbf{W}$) is estimated for simplicity purposes, while the original version of the cognitive architecture of \cite{papadopoulos2020towards} considers the multi-task learning scenario.

\subsection{Constituent entities}
\label{ssec:Entities}

The considered cognitive AI architecture is composed of the following real-world reference entities: a) The robotic platform and b) The human. In order to enable the perception and the collection of critical information about the surrounding environment and the exhibited human behavior, two comprehensive sets of sensing devices are considered; one that can be attached to the robot (denoted $S^r$) and one to the human (denoted $S^h$).
In particular, the set of potentially supported types of sensors for each case, which can be significantly broad and also depends on the particular application scenario, are defined as follows:
\begin{eqnarray}\nonumber
S^r=\{s^r_i | i\in[1,I], i\in\mathbb{N}, I\in\mathbb{N}   \}\\\nonumber
=\{Vision, Light, Temperature, Chemical, Force,\\
Acoustic, Gas, Motion, Pressure, Position, ... \}
\end{eqnarray} 
\begin{eqnarray}\nonumber
S^h=\{s^h_k | k\in[1,K], k\in\mathbb{N}, K\in\mathbb{N}   \}\\\nonumber
=\{Gaze, Heart~rate, Stress~level, Motion,\\
Fatigue, Position, EEG, ... \}
\end{eqnarray} 

Regarding the robotic platform, this refers to the actual mechatronic equipment to be deployed. Depending on the specific operational scenario and the targeted robotic task (that may cover a large set of possible perception, cognition, motor and interaction functionalities), multiple types of robots can be used, supporting different requirements for mobility, positioning, manipulation, communication, size, payload, etc. The set of available types of robots is denoted:
\begin{eqnarray}\nonumber
R=\{r_j | j\in[1,J], j\in\mathbb{N}, J\in\mathbb{N}   \}\\
=\{Arm, AGV, Humanoid, UAV, Vehicle, Industrial, ...   \}
\end{eqnarray} 
Taking into account the above-mentioned formalisms, a robotic platform $P_l$ can be fully specified as follows:
\begin{eqnarray}\label{eq:roboticPlatform}
P_l=\{ S^r_l, R_l | S^r_l \subseteq S^r, R_l \subseteq R \}, l\in[1,L],
\end{eqnarray} 
where $L$ denotes the total number of robotic platforms present in the examined cognitive environment.

With respect to the human factor, although all types of LfD demonstration means methodologies (namely kinesthetic teaching, teleoperation and passive observation, as detailed in Section \ref{sec:Introduction}) are supported by the cognitive AI architecture, the adoption of a teleoperation-based approach is considered to exhibit significant advantageous characteristics. In particular, an intuitive and sophisticated teleoperation scheme is foreseen that is based on the combined used of Augmented Reality (AR) visualization mechanisms and eXplainable AI (XAI) technologies. Regarding the former, AR tools are employed in order to allow the human user to perform a physical inspection of the robot exhibited behaviors (and hence to identify malfunctions, hazardous situations, deviations from desired policies, etc.), while at the same time being constantly provided with key detailed insights about the AI processes being applied (e.g. the AI modules being used, their estimated outputs, how specific decisions are reached, etc.). Concerning XAI methods, such techniques are adopted in order to provide precise explanations/insights to the human operator regarding the deviation of the robot behavior from the desired one, i.e. enabling an in depth inspection of the robot behavior. It needs to be mentioned that the set of human operators involved in the designed cognitive architecture is denoted $H=\{h_m | m\in[1,M], m\in\mathbb{N}, M\in\mathbb{N}\}$.

\subsection{Collective cognitive AI layer}
\label{ssec:LearningEnvironment}

A collective cognitive AI layer operates at the upper level of the designed architecture and its core functionality is based on the fundamental mechanism of the FL paradigm, as detailed in Section \ref{ssec:FLConceptualization}. The main building blocks of the cognitive AI layer, their formalism and detailed explanation of their functionalities are provided in the followings:\\
\underline{{\it{Network nodes}}}: Every robotic platform $P_l$ corresponds to a network node of the defined architecture. Each $P_l$ stores locally the generated data, which largely contain information collected from the set of sensors $S^r_l$ incorporated by $P_l$.\\
\underline{\it{Global AI model}}: The goal of the overall framework is to collectively create a global AI model denoted $\textbf{W} \leftarrow f_1(S^r, R)$, underpinning the implemented cognitive process present in the examined environment, where $f_1(.)$ implies a generalized function or process that defines the exact NN-based materialization of model $\textbf{W}$ while considering $S^r$ and $R$ as input parameters. However, each network node $P_l$ can maintain a local/customized version of $\textbf{W}$.\\
\underline{\it{Learning methodology}}: Regarding the specific methodology to be followed for refining the robot learned policies (i.e. for updating AI model $\textbf{W}$), different options can be investigated (e.g. reinforcement learning, transfer learning, active learning, etc.), as discussed in Section \ref{sec:Introduction}. The most suitable selection depends on the particularities of the application domain and the examined $\textbf{W}$.\\
\underline{\it{Local parameter updates}}: Regardless of the particular learning methodology selected, each robotic platform $P_l$ can estimate updates for model $\textbf{W}$, using its locally stored data. More specifically, the following local parameter update mechanism is applied:
\begin{eqnarray}\label{eq:localModels}
\textbf{W}^{l,m} \leftarrow \textbf{W}^{l,m} - lr^l  \cdot \nabla \mathcal{L} (\textbf{W}^{l,m}, \delta^l),
\end{eqnarray} 
where $\textbf{W}^{l,m}$ is the local/customized version of the global model $\textbf{W}$ with respect to human teacher $h_m$, $lr^l$ is the local learning rate, $\nabla$ denotes the gradient of a function, $\mathcal{L}(.)$ represents the loss function defined for $\textbf{W}$ and $\delta^l$ is the locally stored dataset. 
Consequently, the local parameter updates, which are iteratively estimated, to be sent to the central node (aggregator) are computed as follows:
\begin{eqnarray}\label{eq:localModelUpdate}
\Delta \textbf{W}^{l,m} = \textbf{W}^{l,m} - \textbf{W}
\end{eqnarray} 
{\underline{{\it{User profiling}}}: For efficiently initially interpreting and subsequently incorporating human feedback, in parallel with the $\textbf{W}^{l,m}$ estimation process, an individual user profile $\textbf{Q}^{l,m} \leftarrow f_2(\delta^l, S^h_m)$ is constructed for every human subject $h_m$ at every node $P_l$. The latter is estimated using the appropriate sensorial data $S^h_m$ to model the observed human behavior and a generalized function $f_2(.)$ that defines the exact (NN-based) implementation of the user profile while considering $S^h_m$ as input parameters. Under the current conceptualization, DL-based approaches are considered for generating $\textbf{Q}^{l,m}$, as in \cite{farnadi2018user} and \cite{liang2020drprofiling}, aiming at combining increased modelling capabilities and easier integration to the designed FL-based framework.\\
\underline{\it{Global model update}}: Having computed the local parameter updates $\Delta \textbf{W}^{l,m}$ and the corresponding user profiles $\textbf{Q}^{l,m}$ (using locally generated and processed data $\delta^l$), these are sent to the central node (aggregator) so as to periodically produce updated versions of global model $\textbf{W}$. According to the conventional FL mechanism, an updated version $\textbf{W}'$ of $\textbf{W}$ is generated on a regular basis, by applying a simple averaging operator, as follows:
\begin{eqnarray}\label{eq:FLGoblaModels}
\textbf{W}' = \textbf{W} + lr^g  \cdot \boldsymbol{\Gamma}\\\label{eq:FLGoblaUpdate}
\boldsymbol{\Gamma} = \frac{1}{\Lambda} \sum_{(l,m)} \Delta \textbf{W}^{l,m},
\end{eqnarray} 
where $lr^g$ denotes the global learning rate and $\Lambda$ the total number of received $\Delta \textbf{W}^{l,m}$ updates. For robustly confronting the observed variance in the behavior of the large number of involved human teachers $h_m$, the designed cognitive architecture (apart from possible sensorial data $S^r_l$ pre-processing for invariance incorporation) encompasses the estimated user profiles $\textbf{Q}^{l,m}$ in the global model update process, modifying (\ref{eq:FLGoblaUpdate}) to the following general formalism:
\begin{eqnarray}\label{eq:FLGlobalUpdateHuman}
\boldsymbol{\Gamma} = \Phi (\{\textbf{Q}^{l,m}\},\{ \Delta \textbf{W}^{l,m}\}), 
\end{eqnarray} 
where $\Phi(.)$ denotes a generalized function that combines the available $\textbf{Q}^{l,m}$ and $\Delta \textbf{W}^{l,m}$, estimated at every network node $P_l$. Depending on the particularities of the selected application domain (e.g. supported $\textbf{W}$, $S^r$, $S^h$, $R$, etc.), the following main and purely data-driven materializations of $\Phi(.)$, which are capable of being combined, are considered in \cite{papadopoulos2020towards}: a) User weighting, where each human teacher $h_m$ is modulated by a different weight factor based on a similarity metric from the global user model, b) Parameter weighting, which assigns varying weights to the parameters of $\textbf{W}$, based on the degree of correlation among the parameters of $\textbf{W}^{l,m}$ and $\textbf{Q}^{l,m}$, and c) User clustering, which considers the generation of multiple instances $\textbf{W}_{\eta}$ of the global model $\textbf{W}$, in order to better capture the heterogeneity of data distributions across the different users $h_m$. In the remainder of this work the above-mentioned `User weighting' conceptualization is considered and the focus is put on thoroughly defining an efficient $\textbf{Q}^{l,m}$ formalism (Section \ref{sec:UserProfile}), although `Parameter weighting' and `User clustering' (as well as any combination of the three alternatives) are equally applicable. In particular, the `User weighting' approach adaptively adjusts the contribution of each human teacher $h_m$ by a different weight factor based on his/her exhibited behavior, as follows:
\begin{eqnarray}\nonumber
\boldsymbol{\Gamma} = \frac{1}{E} \sum_{(l,m)} \varepsilon (\textbf{Q}^{l,m}) \cdot \Delta \textbf{W}^{l,m}\\\nonumber
\varepsilon (\textbf{Q}^{l,m})=  1 / \lVert \textbf{Q}^{g} - \textbf{Q}^{l,m} \rVert\\
E = \sum_{(l,m)} \varepsilon (\textbf{Q}^{l,m}),\label{eq:UserWeight}
\end{eqnarray} 
where $\varepsilon (\textbf{Q}^{l,m})$ denotes the weight factor for each $h_m$ at every $P_l$, $\textbf{Q}^{g}$ the corresponding global user model (e.g. estimated through the same FL-based mechanism used for constructing $\textbf{W}$) and $\lVert . \rVert$ a dissimilarity score metric (e.g. Euclidean distance between the parameters of the involved user models).

\begin{figure}[t]
\centering
\includegraphics[width=0.45\textwidth]{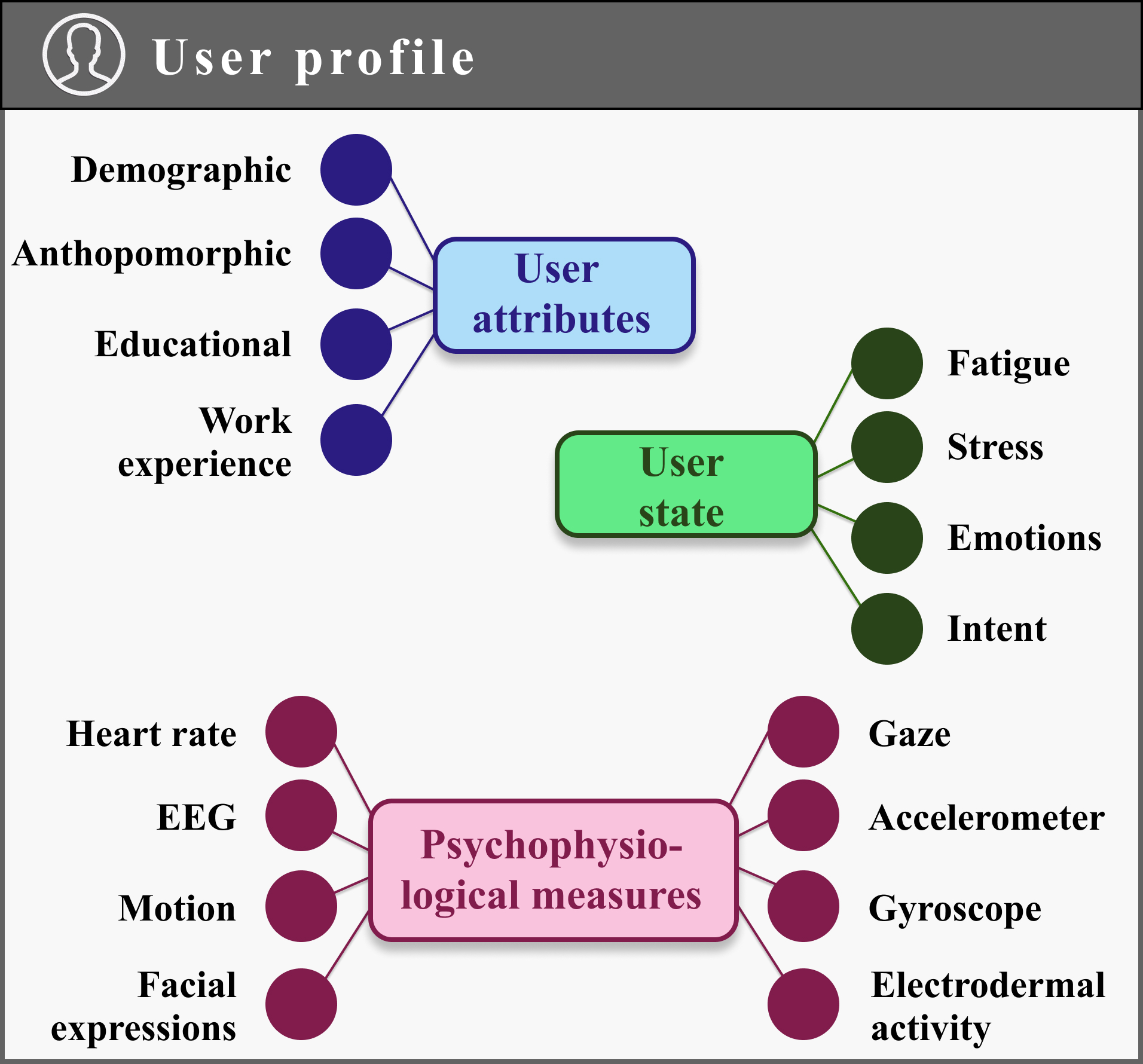}
\caption{Hierarchy of information sources for generating user profile $\textbf{Q}^{l,m}$.}
\label{f:UserModelData}
\end{figure}

\section{DL-based user modelling and knowledge aggregation}
\label{sec:UserProfile}

Incorporating human feedback constitutes a fundamental part of the LfD approach and, consequently, of the employed cognitive architecture. However, the latter poses additional challenges to the problem formulation that need to be efficiently addressed (e.g. variance in human performance, variability in knowledge across human subjects, learning from noisy/imprecise human input, possible conflicts in provided human feedback information, handling of individuals with different idiosyncrasies, learning from very large or very sparse datasets, incremental learning, etc.). On the other hand, the strength of the utilized FL-based mechanism (detailed in Section \ref{sec:FL}) lies on incorporating a very large number $\Lambda$ of samples and multiple iterative updates of $\textbf{W}'$ that will likely lead to the convergence to well-performing and robust $\textbf{W}$ models, while the network nodes $P_l$ contributing in (\ref{eq:FLGoblaUpdate}) may be sampled out of the available ones (usually in a random way). However, combining information ($\Delta \textbf{W}^{l,m}$) related to different human subjects $h_m$ (that presumably exhibit highly diverse and varying behavior) is in turn very likely to lead the FL process to be confined to a local maximum or even to a non-convergence of the FL procedure; hence, jeopardising the overall learning process. To this end, defining a detailed and robust user profile model $\textbf{Q}^{l,m}$ is of vital importance for enabling: a) The accurate assessment and interpretation of the demonstrated behavior of each human teacher $h_m$ and, subsequently, b) The appropriate adjustment of the importance of the provided human feedback from the same individual or across multiple ones, during the knowledge aggregation step.

\begin{figure*}[t]
\centering
\includegraphics[width=0.45\textwidth]{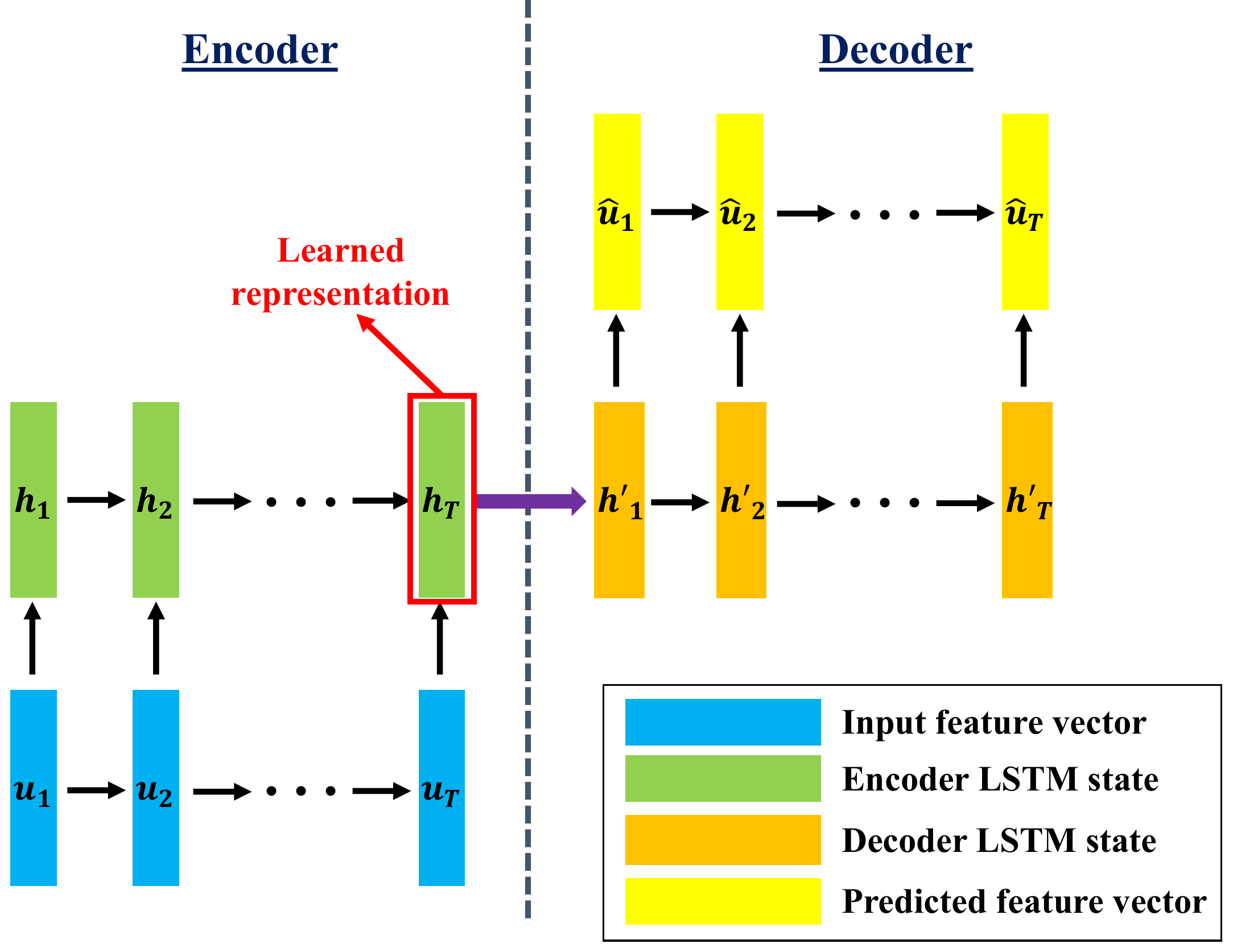}
\caption{Employed LSTM autoencoder for producing psychophysiological measurements representation ${\bf{DR}}^{l,m}_{\nu,\rho}$.}
\label{f:LSTMAutoencoder}
\end{figure*}

In general, the purpose of defining user profile formalisms is to cover physical (e.g. human actions, physical capabilities, etc.), cognitive (e.g. intention, personality, etc.) and social (e.g. non-verbal cues, emotions, etc.) aspects, in order to model and efficiently interpret the exhibited human activity \cite{rossi2017user}. So far, the problem has been investigated and assessed from different disciplines and scientific perspectives, while methodologies that rely on the use of Machine Learning (ML) techniques and the integration of the generated models in broader computational/cognitive systems continuously receives increasing attention. One of the most popular examples of the latter category of methods constitutes the so called recommender systems (that aim to predict the preferences of each individual human user) \cite{mu2018survey}\cite{zhang2019deep} and especially the so-called `collaborative filtering' class \cite{ji2020dual}. However, the introduced profiles so far model relatively simple types of human behavior aspects (e.g. correlations of individual users and different information items \cite{he2017neural}).

\subsection{Considered information sources}
\label{ssec:InformationSources}

Under the proposed conceptualization, a fine-grained multi-level DL-based computational approach is introduced for modelling user profile $\textbf{Q}^{l,m}$, grounded on the efficient processing and sophisticated combination of various heterogeneous and diverse information streams. The overall goal is to meet the demanding requirements of the LfD setting for modelling the exhibited user behavior, focusing on the following two main pillars: a) Accurately interpreting the captured human behavior both in the short- and the long-term cases and b) Effectively handling the variance of cross-subject (and often conflicting) information and its efficient merging. For that purpose, the following (hierarchically organized) types of information sources are considered for generating $\textbf{Q}^{l,m}$ (as also illustrated in Fig. \ref{f:UserModelData}):
\begin{itemize}
\item {\it{User attributes}}: These include valuable and explicitly defined information concerning each human teacher's $h_m$ background, covering aspects like demographic (e.g. age, sex, race, origin, etc.), anthropomorphic (e.g. height, weight, possible disabilities, etc.), educational (e.g. level and types of education, etc.), work experience (e.g. duration, responsibilities/tasks, etc.), etc.;
\item {\it{User state}}: This corresponds to particular modules/pipelines that (automatically) detect specific human behavior characteristics, e.g. fatigue detection, stress detection, emotion recognition, intention prediction, etc. For their extraction sensorial data types belonging to $S^r$ and $S^h$ are used;
\item {\it{Psychophysiological measurements}}: These comprise the actual raw sensorial data information streams captured (supporting data types of both $S^r$ and $S^h$ sets) and convey critical details about the user's mental state, personality and idiosyncrasy, e.g. heart rate, electroencephalogram (EEG), motion, facial expressions, gaze, accelerometer, gyroscope, electrodermal activity, etc.
\end{itemize}
It needs to be highlighted that the specific information sources (belonging to each of the above mentioned categories) to be considered depends on the particularities of the selected application case.

\subsection{Processing of collected data}
\label{ssec:DataProcessing}

According to the formalism of the cognitive architecture for LfD learning defined in Section \ref{sec:FL}, each human teacher $h_m$ can provide, in relation to a given robotic platform $P_l$, feedback information (denoted $\mu^{l,m}_\nu$) regarding the robot exhibited behavior, where $\nu$ ($\nu\in[1,N], \nu\in\mathbb{N}, N\in\mathbb{N}$) indicates the index of the particular feedback session (denoted $\xi^{l,m}_\nu$) that consists of $T_\nu$ time steps. The user attributes provided for human $h_m$, after being appropriately numerically encoded, are concatenated to form static feature vector ${\bf{DE}}^m$. On the other hand, the outputs of the specific feature detectors (that summarize the user's state) for human $h_m$ at node $P_l$ during session $\xi^{l,m}_\nu$, after also being appropriately numerically encoded (if needed), are similarly concatenated to produce static feature vector ${\bf{DS}}^{l,m}_\nu$. With respect to the corresponding psychophysiological measurements captured during session $\xi^{l,m}_\nu$, an individual multi-dimensional observation sequence $OS^{l,m}_{\nu,\rho}$ (of length $T_\nu$ samples) is formed for every considered sensor type implied by index $\rho$. In order to estimate a static feature vector ${\bf{DR}}^{l,m}_{\nu,\rho}$ (out of each $OS^{l,m}_{\nu,\rho}$ sequence), a DL-based unsupervised feature representation learning methodology is considered. Specifically, an individual LSTM autoencoder, similar to the one described in \cite{srivastava2015unsupervised} and whose architecture is illustrated in Fig. \ref{f:LSTMAutoencoder}, is introduced for each $OS^{l,m}_{\nu,\rho}$, since every $OS^{l,m}_{\nu,\rho}$ is in general related to a different sampling frequency and, hence, consists of a different number of samples. As can be seen in Fig. \ref{f:LSTMAutoencoder}, the autoencoder consists of two LSTM models, namely an encoder and a decoder one. The encoder receives as input a series of feature vectors (i.e. ${\bf{u}}_1$, ${\bf{u}}_2$, ... ${\bf{u}}_T$) and updates at each time step its vectorial fixed-length internal state representation (i.e. ${\bf{h}}_1$, ${\bf{h}}_2$, ... ${\bf{h}}_T$), which models/encodes the input sequence until that time step. During the decoding phase, the second LSTM is initialized with internal state ${\bf{h}}'_1 \equiv {\bf{h}}_T$ and, through its fundamental sequential modelling mechanism, it recursively estimates new internal states (i.e. ${\bf{h}}'_1$, ${\bf{h}}'_2$, ... ${\bf{h}}'_T$) that correspondingly produce a respective series of prediction feature vectors (i.e. ${\bf{\widehat{u}}}_1$, ${\bf{\widehat{u}}}_2$, ... ${\bf{\widehat{u}}}_T$). The overall goal of the autoencoder is to adjust its internal parameters so that sequences ${\bf{u}}_1$, ${\bf{u}}_2$, ... ${\bf{u}}_T$ and ${\bf{\widehat{u}}}_1$, ${\bf{\widehat{u}}}_2$, ... ${\bf{\widehat{u}}}_T$ to be as similar as possible. Vector ${\bf{h}}_T$ that characterizes the overall input sequence ${\bf{u}}_1$, ${\bf{u}}_2$, ... ${\bf{u}}_T$ is considered to coincide with ${\bf{DR}}^{l,m}_{\nu,\rho}$ defined above.

\subsection{Modeling of user profile}
\label{ssec:UserProfileModeling}

The ultimate goal of designing a model of the human user profile $\textbf{Q}^{l,m}$ (Section \ref{sec:FL}) is to provide a computational scheme that will simultaneously analyse, cross-correlate and eventually fuse the different heterogeneous information streams for producing a joint/compact representation that will efficiently encode/express the salient characteristics of each human teacher $h_m$ (e.g. mental state, idiosyncrasy, preferences, etc.) in an implicit way. The latter is of paramount importance for interpreting the human behavior during the knowledge aggregation step, as will be discussed in Section \ref{ssec:KnowledgeAggregation}.

For creating a robust user profile, a DL-based hierarchical multi-stream unsupervised feature fusion approach is adopted. In particular, a multi-stream stacked autoencoder, similar to the one described in \cite{gao2020survey} and whose general architecture is illustrated in Fig. \ref{f:StackedAutoencoder}, is introduced. Each of the $3$ defined autoencoder streams is in direct accordance to the information sources definition of Section \ref{ssec:InformationSources}. In particular, a stacked autoencoder is a neural network consisting of several layers of sparse autoencoders, where the output of each hidden layer is connected to the input of the successive hidden layer. The designed autoencoder considers a multi-stream architecture, which aims at initially capturing the correlations among the features of the same stream (while also leveraging knowledge from the other streams) and subsequently modelling the inter-dependencies among the different streams, as can been seen in Fig. \ref{f:StackedAutoencoder}. Similarly to the LSTM autoencoder (Section \ref{ssec:DataProcessing}), the goal during the training process is for the input and output feature vectors to be as similar as possible. Moreover, the learned multi-modal representation (feature vector) of the middle/internal encoding layer compactly summarizes the salient attributes of the input feature vectors.

\begin{figure*}[t]
\centering
\includegraphics[width=0.60\textwidth]{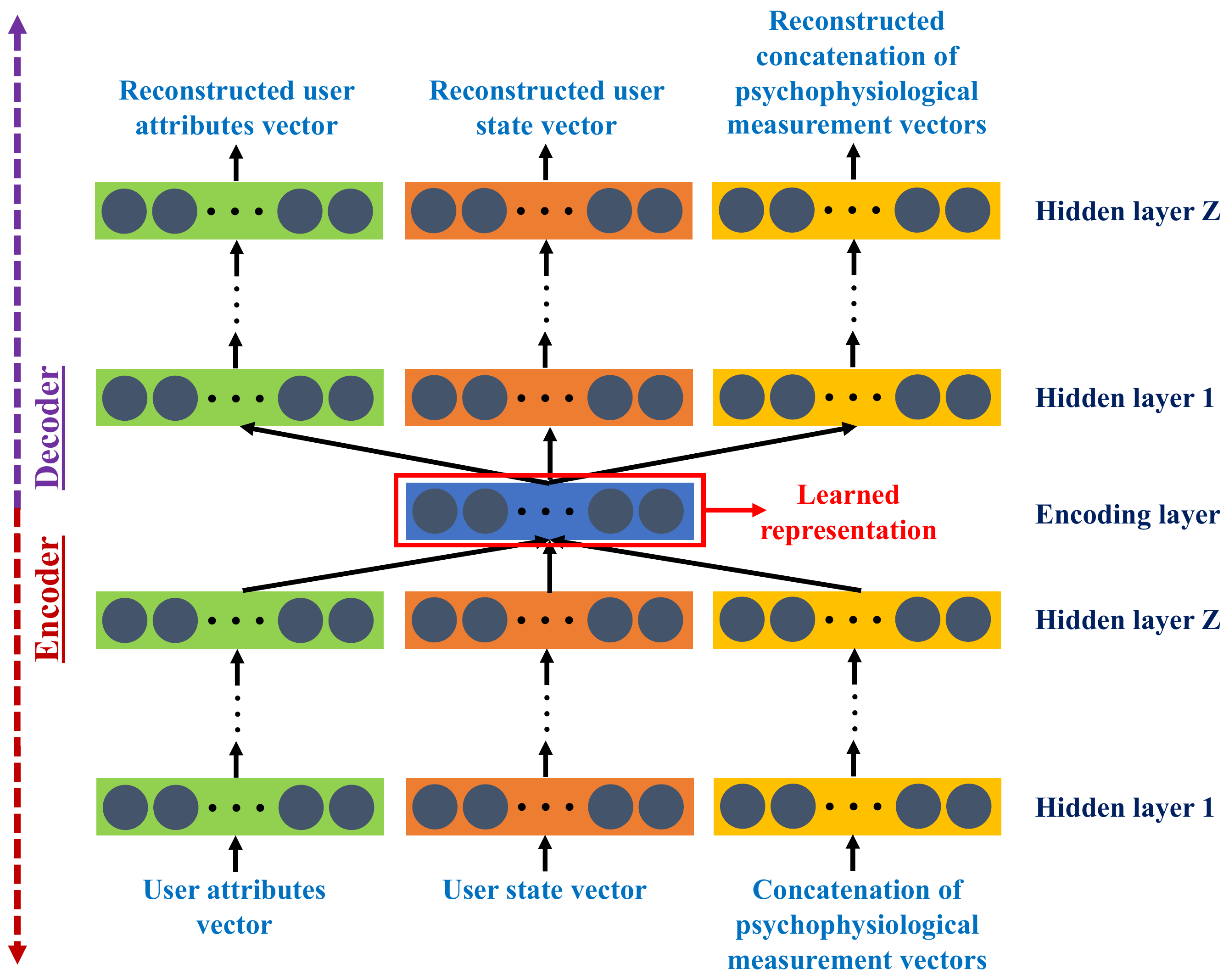}
\caption{Designed stacked autoencoder for generating user profile model $\textbf{Q}^{l,m}$.}
\label{f:StackedAutoencoder}
\end{figure*}

Regarding the actual user profile generation process, an individual stacked autoencoder is considered with respect to every $\textbf{Q}^{l,m}$. The autoencoder receives as input (with respect to each associated feedback session $\xi^{l,m}_\nu$) the following feature vectors: a) User attributes vector ${\bf{DE}}^m$, b) User state vector ${\bf{DS}}^{l,m}_\nu$, and c) Concatenation of psychophysiological measurement vectors ${\bf{DR}}^{l,m}_{\nu,\rho}$. The estimated state representation of the encoding layer (denoted ${\bf{HS}}^{l,m}_\nu$) comprises a static feature vector that summarizes the salient behavioral characteristics of human teacher $h_m$ at node $P_l$ during session $\xi^{l,m}_\nu$. Moreover, the parameters of the learned stacked autoencoder (or practically only those of the encoder part) constitute the actual user profile model $\textbf{Q}^{l,m}$ introduced in the FL-based formalization of Section \ref{sec:FL}.

\subsection{Knowledge aggregation}
\label{ssec:KnowledgeAggregation}

Having discussed the development of detailed user behavior profiles (Section \ref{ssec:UserProfileModeling}), the challenging issue of analysing and aggregating human knowledge (in the form of feedback demonstrations in the LfD setting) is investigated in this section. The overall conceptualization concentrates on two main axes with respect to the interpretation of the exhibited human behavior: a) Short-term analysis for combining information cues originating from the same human teacher and b) Long-term analysis for integrating knowledge derived from different individuals.

Regarding the short-term analysis, this targets to in principle interpret the observed human behavior when providing feedback information $\mu^{l,m}_\nu$ during session $\xi^{l,m}_\nu$. The goal is to adaptively adjust the importance of each individual feedback sample $\mu^{l,m}_\nu$ when updating the local/personalized model $\textbf{W}^{l,m}$ for human teacher $h_m$ at node $P_l$, by taking into account user profile related information estimated during each session $\xi^{l,m}_\nu$. For achieving that, the loss function $\mathcal{L} (\textbf{W}^{l,m}, \delta^l)$ defined in (\ref{eq:localModels}), which is used for updating local model $\textbf{W}^{l,m}$, is formalized as follows:

\begin{eqnarray}\nonumber
\mathcal{L} (\textbf{W}^{l,m}, \delta^l) = \frac{\sum_\nu  \omega^{l,m}_\nu \cdot \mathcal{L}_\nu (\mu^{l,m}_\nu, \textbf{W}^{l,m})}{\sum_\nu  \omega^{l,m}_\nu}\\ \label{eq:shortTermAnalysis}
\omega^{l,m}_\nu =  1 / \lVert {\bf{HS}}^{l,m}_\nu - {\overline{\bf{HS}}}^{l,m}_\nu \rVert, 
\end{eqnarray} 
where $\mathcal{L}_\nu (\mu^{l,m}_\nu, \textbf{W}^{l,m})$ is the loss factor for sample $\mu^{l,m}_\nu$, $\omega^{l,m}_\nu$ its associated weight factor, ${\overline{\bf{HS}}}^{l,m}_\nu$ the mean value of ${\bf{HS}}^{l,m}_\nu$ (calculated while considering all available $\mu^{l,m}_\nu$ samples) and $\lVert . \rVert$ a dissimilarity score metric (e.g. Euclidean distance). $\omega^{l,m}_\nu$ assigns varying importance to each $\mu^{l,m}_\nu$, contrary to the conventional practice of associating all $\mu^{l,m}_\nu$ samples with equal weight (i.e. $\omega^{l,m}_\nu=1$). The motivation behind the latter choice lies on the fundamental consideration that observed human behaviors ${\bf{HS}}^{l,m}_\nu$ that deviate significantly from the expected personalized norm ${\overline{\bf{HS}}}^{l,m}_\nu$ (i.e. outliers) should receive decreased importance, since it is assumed that they are likely to be caused by undesirable user profile states (e.g. when in extreme situations of fatigue, stress, decreased attention, etc.).

With respect to the long-term analysis, the stacked autoencoder computational model $\textbf{Q}^{l,m}$ described in Section \ref{ssec:UserProfileModeling} is used in the `User weighting' scheme defined in (\ref{eq:UserWeight}), where $\textbf{Q}^{l,m}$ that are similar to the respective global $\textbf{Q}^{g}$ user profile receive increased impact. Similarly to the above, the explanation of the latter states that model updates $\Delta \textbf{W}^{l,m}$ ((\ref{eq:localModelUpdate})) should receive greater importance proportionally to the associated user profiles $\textbf{Q}^{l,m}$ being similar to the global norm or expected behavior $\textbf{Q}^{g}$. In other words, human teacher behaviors ($\textbf{Q}^{l,m}$) that deviate significantly from the global norm ($\textbf{Q}^{g}$) are considered to be due to frequent undesirable user profile states, which could very likely in turn jeopardize the convergence of the overall FL-based learning scheme; hence, decreasing their impact would facilitate and reinforce the robustness of the overall LfD learning environment. It needs to be highlighted that in the original work of \cite{papadopoulos2020towards} no specific materialization of $\textbf{Q}^{l,m}$ is provided.

\section{Conclusions}
\label{sec:Conclusions}
In this paper, the recently introduced Federated Learning (FL)-based conceptualization for developing large-scale multi-agent Learning from Demonstration (LfD) human-robot collaborative environments was further elaborated and enhanced, by incorporating a novel Deep Learning (DL)-based user profile formulation for estimating a fine-grained representation of the exhibited human behavior. The overall integrated scheme supports both short- and long-term analysis/interpretation of the observed human teacher behavior, in order to adaptively adjust the importance of the collected feedback samples when aggregating information originating from the same and different human individuals, respectively. Future work includes the implementation and evaluation of the designed framework in real-world settings, the development of more sophisticated user profile models and the incorporation of more elaborate DL formalisms (e.g. (self-) attention schemes, transformer networks, etc.).

\section*{ACKNOWLEDGMENTS}
The work presented in this paper was supported by the ICS-FORTH internal RTD Programme `Ambient Intelligence and Smart Environments'.

\bibliographystyle{IEEEtran}
\bibliography{ROMAN_v2}

\end{document}